\documentclass[10pt,twocolumn,letterpaper]{article}

\usepackage{iccv}              
\usepackage[dvipsnames]{xcolor}
\usepackage{etoolbox}
\usepackage{longtable}
\usepackage{pdflscape}
\usepackage{graphicx}
\usepackage{siunitx}
\usepackage{amsmath}
\usepackage{amssymb}
\usepackage{nicefrac}
\usepackage[accsupp]{axessibility}
\usepackage{colortbl}%

\usepackage{booktabs}

\usepackage{graphicx}
\usepackage[ruled,vlined]{algorithm2e}

\usepackage{inconsolata}
\usepackage{multirow}
\usepackage{tabularx}
\usepackage{colortbl}
\usepackage{pifont}
\usepackage{multicol}
\usepackage{booktabs}
\usepackage{makecell}

\usepackage{amssymb}
\usepackage{adjustbox}
\usepackage{enumerate}
\usepackage{amsmath}

\usepackage{placeins}

\definecolor{myred}{HTML}{C00000}
\definecolor{myorange}{HTML}{ED7D31}
\definecolor{mygreen}{HTML}{548235}
\definecolor{myblue}{HTML}{00B0F0}
\definecolor{mypurple}{HTML}{7030A0}

\definecolor{mydarkgreen}{rgb}{0.0, 0.5, 0.0}

\usepackage{marvosym}
\usepackage{utfsym}
\usepackage{xspace}
\usepackage{pifont}
\usepackage{arydshln}
\usepackage[T1]{fontenc}
\definecolor{LGray}{gray}{0.97}

\usepackage{sidecap}
\usepackage{booktabs}
\renewcommand{\arraystretch}{1}
\usepackage{afterpage}

\usepackage[dvipsnames]{xcolor}
\usepackage{bbm}
\usepackage{stfloats}

\newcommand{\ourmethod}{AutoOcc}

\newif\ifdrafting
\draftingtrue 
\ifdrafting
    \newcommand{\todo}[1]{{\leavevmode\color[rgb]{1,0,0}[TODO: #1]}}
    
    \definecolor{tabfirst}{rgb}{1, 0.7, 0.7} 
    \definecolor{tabsecond}{rgb}{1, 0.85, 0.7} 
    \definecolor{tabthird}{rgb}{1, 1, 0.7} 
    \definecolor{tabzero}{rgb}{0.95, 0.95, 0.95} 
\else
    \newcommand{\todo}[1]{}
    \newcommand{\ds}[1]{}
    \newcommand{\mh}[1]{}
    
\fi

\newcommand{\aftertab}{\vspace{-1em}}
\newcommand{\afterfig}{\vspace{-1.25em}}

\definecolor{iccvblue}{rgb}{0.21,0.49,0.74}
\usepackage[pagebackref,breaklinks,colorlinks,allcolors=iccvblue]{hyperref}

\def\paperID{2989} 
\def\confName{ICCV}
\def\confYear{2025}

\title{\textcolor{TealBlue}{AutoOcc}: \textcolor{TealBlue}{Auto}matic Open-Ended Semantic \textcolor{TealBlue}{Occ}upancy Annotation via Vision-Language Guided Gaussian Splatting}

\author{
Xiaoyu Zhou\textsuperscript{1}
~ Jingqi Wang\textsuperscript{1}
~ Yongtao Wang\textsuperscript{1}\thanks{Corresponding author.}
~ Yufei Wei\textsuperscript{2} \\
~ Nan Dong\textsuperscript{2}
~ Ming-Hsuan Yang\textsuperscript{3} \\
{\textsuperscript{1}Wangxuan Institute of Computer Technology, Peking University} \\ 
~ {\textsuperscript{2}Chongqing Changan Automobile Co., Ltd}
~ {\textsuperscript{3}University of California, Merced} \\
}

\begin{document}
\maketitle
\begin{abstract}
Obtaining high-quality 3D semantic occupancy from raw sensor data remains an essential yet challenging task, often requiring extensive manual labeling. In this work, we propose \textbf{\ourmethod{}}, a vision-centric \textbf{auto}mated pipeline for open-ended semantic \textbf{occ}upancy annotation that integrates differentiable Gaussian splatting guided by vision-language models. We formulate the open-ended semantic 3D occupancy reconstruction task to automatically generate scene occupancy by combining attention maps from vision-language models and foundation vision models. We devise semantic-aware Gaussians as intermediate geometric descriptors and propose a cumulative Gaussian-to-voxel splatting algorithm that enables effective and efficient occupancy annotation. Our framework outperforms existing automated occupancy annotation methods without human labels. \ourmethod{} also enables open-ended semantic occupancy auto-labeling, achieving robust performance in both static and dynamically complex scenarios.
\end{abstract}    
\section{Introduction}
\label{sec:intro}
3D semantic occupancy has attracted a considerable amount of attention in autonomous driving~\cite{wang2021learning, tong2023scene, wang2024panoocc} and embodied intelligence~\cite{ramakrishnan2020occupancy, ramakrishnan2021exploration, chaplot2021seal}, demonstrating great potential to facilitate understanding of 3D scenes and perception of irregular objects. Despite its promising applications, automatic generation of precise and complete semantic occupancy annotations from raw sensor data remains a fundamental challenge, particularly in the pursuit of cost-effective solutions for real-world deployment.

Vision-centric automated 3D semantic occupancy annotation has long been undervalued, while existing occupancy annotation pipelines heavily rely on LiDAR point clouds (Table~\ref{tab:labeling}), requiring human pre-annotations and labor-intensive post-processing (over 4k+ human hours for nuScenes~\cite{tong2023scene}).
Current automated or semi-automated annotation pipelines primarily follow three paths. 
(1) Automated-assisted manual annotation, which is labor-intensive and costly. 
(2) Point cloud voxelization guided by manual annotation priors relies heavily on manual priors and multi-stage post-processing, making it time-consuming. 
(3) 2D-to-3D projection-based methods, which simply merge 2D segmentation results into 3D point clouds or meshes, struggle to ensure precise 3D consistency.
These annotation methods heavily rely on LiDAR point clouds while overlooking semantic and geometric cues from multi-view images. Given that LiDAR point clouds are inherently sparse and incomplete, they are insufficient for comprehensive scene modeling.
These approaches also employ voxel-based scene representations that require excessive parameters and incur redundant computational costs.
Recent self-supervised occupancy models~\cite{huang2024selfocc, gan2024comprehensive, zhang2023occnerf, boeder2024occflownet, wan2024gaussianocc} have eliminated the need for extensive labeled training data by leveraging 2D features from image inputs and semantic information from visual foundation models (VFMs), such as SAM~\cite{kirillov2023segment} and OpenSeed~\cite{zhang2023simple}. Nevertheless, these methods struggle to ensure complete, consistent scene occupancy, and exhibit limited generalization across diverse scenes. 

Additionally, these pipelines are all confined to closed-set or open-set occupancy classes that require predefined categories. However, real-world scenes often involve open-ended occupancy—objects outside any predefined category, making it unwise to label all undefined semantics as ``others.'' For example, self-driving vehicles may encounter collapsed poles or plastic sheets on road surfaces that require distinct occupancy annotations for safe driving strategies.


\begin{table*}[ht]
\footnotesize
    \renewcommand\arraystretch{1.2}
    \setlength{\tabcolsep}{0.015\linewidth}
    \centering
    \caption{\textbf{Comparisons between \ourmethod{} and existing semantic occupancy annotation pipelines.} The definitions of closed-set, open-set, and open-ended are introduced in Section~\ref{open}. Our method achieves high-quality occupancy annotation without additional manual labeling or post-processing while maintaining superior speed and generalization. C represents camera, and L denotes LiDAR.
    }
    \label{tab:labeling_comparison}
    \vspace{-3mm}
    \begin{tabular}{l|ccccccccc}
    \Xhline{0.75pt}
    Method & Categories
 & Modality
 & Manual-label
 & Post-processing
 & Speed
 & Zero-shot
 & Dynamic
\\
        \hline

    Point-based Voxelization~\cite{wei2023surroundocc, Wang_2023_ICCV,tong2023scene} & Close-set & L & 3D GT & Human & Slow & \textcolor{red}{\usym{2717}} & \textcolor{ForestGreen}{\usym{2713}} \\
    2D-to-3D Projection~\cite{lu2023ovir, zhang2023sam3d} & Close/Open-set & C\&L & 2D GT & Auto\&Human & Slow & \textcolor{red}{\usym{2717}} & \textcolor{red}{\usym{2717}} \\
    \rowcolor{violet!10} \textbf{Ours (AutoOcc)} & Open-ended & C or C\&L & N/A & N/A & Fast & \textcolor{ForestGreen}{\usym{2713}} & \textcolor{ForestGreen}{\usym{2713}} \\
    \Xhline{0.75pt}
    \end{tabular}
    \label{tab:labeling}
    \vspace{-2mm}
\end{table*}


To address these limitations, we present \ourmethod{}, a fully automated framework for open-ended semantic occupancy annotation that requires neither manual labeling nor predefined categories.
To achieve open-ended semantic occupancy labeling, we employ semantic attention maps generated by vision-language models (VLMs) to describe the scene, constructing a continuously evolving semantic query list. The generated attention maps are used simultaneously to prompt segmentation in SAM and guide instance-level depth estimation from UniDepth, thereby eliminating the need for manual annotations. We further introduce a self-estimated flow module to identify and manage dynamic objects in temporal rendering. We further propose Gaussian Splatting with open-ended semantic awareness (VL-GS) as an intermediate representation, offering a more comprehensive modeling, improved spatiotemporal consistency, and finer geometry with fewer primitives. Compared to densified point clouds and voxels, VL-GS achieves higher representation efficiency, greater accuracy, and reduced memory consumption. The semantic occupancy annotation is then automatically generated end-to-end through cumulative Gaussian-to-voxel splatting.
Extensive experiments demonstrate that \ourmethod{} outperforms existing automated occupancy annotation methods. Our method further exhibits excellent open-ended and zero-shot generalization capabilities, as evidenced by cross-dataset experiments.
Our main contributions include:
\begin{itemize}
\item We present \ourmethod{}, a vision-centric automatic annotation pipeline that supports open-ended semantic 3D occupancy label generation, based on vision-language guided differentiable reconstruction.

\item We devise VL-GS, an efficient and comprehensive scene representation for semantic occupancy annotation. By integrating vision-language attention with visual foundation models, VL-GS effectively handles dynamic objects over time while enhancing both spatiotemporal consistency and 3D geometric detail in semantic occupancy.

\item \ourmethod{} gains notable improvements over the existing automatic occupancy annotation pipelines, even without relying on manual priors. Our method also demonstrates strong generalization and open-ended understanding capabilities across unseen categories and diverse scenes.

\end{itemize}

\section{Related Work}
\label{sec:formatting}

\begin{figure*}[htbp]
  \centering
  \includegraphics[width=.9\linewidth]{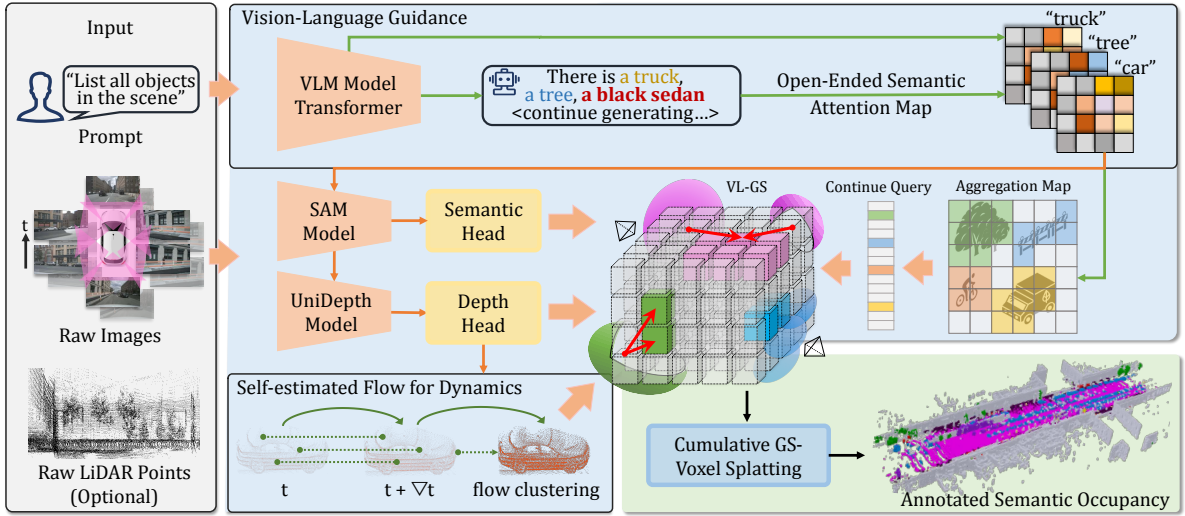}
  \caption{\textbf{Overall pipeline of our method.} \ourmethod{} is a vision-centric automated pipeline for semantic occupancy annotation. Our method starts with multi-view image inputs (optionally with LiDAR), extracts semantic attention maps from VLMs, and refines a dynamic semantic query list. We then propose Vision-Language Guided Gaussian Splatting (VL-GS), incorporating semantic-aware scalable Gaussians and self-estimated flow for dynamic objects. The final occupancy annotation is generated through a forward-pass Cumulative GS-Voxel Splatting. \ourmethod{} demonstrates strong generalization and open-ended annotation capabilities without relying on manual priors or LiDAR.
  }
  \label{fig:overview}
  \afterfig
\end{figure*}

\noindent {\bf{Semantic Occupancy Annotation.}}
Semantic occupancy annotation aims to label semantic 0-1 occupancy from sensor data. However, current automated and semi-automated methods~\cite{wei2023surroundocc, Wang_2023_ICCV, tong2023scene} heavily rely on LiDAR point clouds and human pre-annotated 2D or 3D ground truth. Most of these methods also require time-consuming post-processing and expensive manual purification. In contrast, we design a vision-centric fully automated occupancy annotation pipeline that eliminates the reliance on LiDAR. Our method also integrates VLMs and VFMs, supporting open-ended semantic category annotation.

\noindent {\bf{3D Occupancy Estimation.}}
Semantic 3D occupancy estimation~\cite{tian2024occ3d, zhang2023occformer, Wang_2023_ICCV} aims to estimate the occupancy states and semantics of complex scenes, which is crucial for 3D perception and planning.
Existing learning-based occupancy models~\cite{li2022bevformer, shi2024occupancy, zhang2023occformer, huang2023tri, li2023fb, pan2023uniocc, yu2023flashocc} are heavily reliant on the extensive labeled training data generated by annotation pipelines.
Recent advances in self-supervised methods~\cite{huang2024selfocc, zhang2023occnerf, boeder2024occflownet, wan2024gaussianocc, boeder2024langocc} for estimating 3D occupancy have diminished reliance on costly annotations and can be regarded as online occupancy labeling techniques.
However, these approaches introduce ambiguity and illusions, resulting in misaligned geometry and temporal inconsistencies due to their limited awareness of intricate spatial structures and dynamic objects. They are also hampered by limited cross-dataset and scene-aware generalization capabilities.

To address these limitations, we propose a reconstruction-based occupancy annotation framework that requires no manual 2D or 3D annotations, achieving high-precision open-ended understanding, zero-shot learning, and cross-dataset generalization.

\vspace{1mm}
\noindent {\bf{Scene Representation and Reconstruction.}}
Efficient scene representation is the core of occupancy annotation.
Dense voxel-based methods~\cite{li2022unifying, chen2023voxelnext, li2022voxel, cao2022monoscene} assign each voxel a feature vector, inevitably suffering from high computational cost due to redundant grids. 
As a compressed representation, BEV~\cite{harley2023simple, man2023bev, chambon2024pointbev, li2024fast} encodes 3D information on the ground plane, but struggles to capture diverse 3D geometry using flattened vectors.
By implicitly modeling 3D space, ~\cite{zhang2023occnerf, huang2024selfocc, gan2024comprehensive, pan2024renderocc} create a NeRF-style 3D volume to estimate scene occupancy. 
However, the continuous implicit neural fields struggle with modeling complex dynamic scenes, and dense sampling leads to redundant, memory-intensive operations.
Most recently, 3D Gaussian splatting (3DGS)~\cite{kerbl20233d, yu2024mip, qin2024langsplat} has demonstrated its powerful capability in reconstruction, even for driving scenes~\cite{zhou2024drivinggaussian, tian2024drivingforward, fischer2024dynamic}. 
By treating each vertex as a Gaussian, ~\cite{wan2024gaussianocc} adopts a self-supervised approach for occupancy estimation but results in a dramatic increase in computational cost.

In contrast to prior art, we propose VL-GS, specifically designed to reconstruct semantic instances and dynamic objects, leveraging semantic attention clues from vision-language models. As a more efficient representation, VL-GS achieves high precision and versatile occupancy annotation with reduced cost.

\noindent {\bf Open-World Understanding.} 
\label{open}
Existing open-world understanding methods~\cite{wu2024towards} are confined to 2D images and can be broadly classified into two types: open-set~\cite{scheirer2012toward} and open-ended~\cite{lin2024generative}. Open-set methods~\cite{li2022grounded, liu2024grounding, cheng2024yolo} focus on text-image embedding matching using a predefined vocabulary bank. In contrast, open-ended methods~\cite{lin2024training, lin2024training} continuously update observed object categories via language models. The key difference lies in the reliance on predefined categories, which allows open-ended approaches to produce more precise and comprehensive semantic representations, ultimately enhancing semantic occupancy annotation in open-world scenarios.

\vspace{1mm}
\noindent {\bf VLMs and VFMs.}
Vision language models (VLM)~\cite{kuo2022f, xu2021vlm} and visual foundation models (VFMs)~\cite{ravi2024sam, scheirer2012toward} have shown promising results and generalization ability in various visual tasks.
However, their application to 3D occupancy annotation has received limited attention.
Unlike direct training with 3D annotations, existing foundational vision models~\cite{kirillov2023segment, ke2024segment, liu2023grounding, ren2024grounded} are primarily trained on 2D images, which may challenge the consistency of 3D occupancy across different cameras and frames.
In this work, we explore the potential of applying VLMs~\cite{kirillov2023segment, liu2023grounding, ren2024grounded} to occupancy annotation.

\definecolor{nbarrier}{RGB}{255, 120, 50}
\definecolor{nbicycle}{RGB}{255, 192, 203}
\definecolor{nbus}{RGB}{255, 255, 0}
\definecolor{ncar}{RGB}{0, 150, 245}
\definecolor{nconstruct}{RGB}{0, 255, 255}
\definecolor{nmotor}{RGB}{200, 180, 0}
\definecolor{npedestrian}{RGB}{255, 0, 0}
\definecolor{ntraffic}{RGB}{255, 240, 150}
\definecolor{ntrailer}{RGB}{135, 60, 0}
\definecolor{ntruck}{RGB}{160, 32, 240}
\definecolor{ndriveable}{RGB}{255, 0, 255}
\definecolor{nother}{RGB}{139, 137, 137}
\definecolor{nsidewalk}{RGB}{75, 0, 75}
\definecolor{nterrain}{RGB}{150, 240, 80}
\definecolor{nmanmade}{RGB}{213, 213, 213}
\definecolor{nvegetation}{RGB}{0, 175, 0}
\definecolor{nfence}{RGB}{35, 135, 230}
\definecolor{ntrunk}{RGB}{195,85,85}
\definecolor{npole}{RGB}{213, 0, 139}

\section{Method}
\label{methodd}

As shown in Figure~\ref{fig:overview}, we provide an overview of our proposed auto-annotation pipeline. Given a multi-view image sequence as input, we employ a fixed text prompt to enumerate all possible objects within the scene. Concurrently, our method supports LiDAR input, serving as a robust geometric prior constraint.

\subsection{Vision-Language Guidance}
Human annotations are both costly and labor-intensive. In contrast, world prior knowledge acquired from Vision-Language Models (VLMs) offers a cost-effective and efficient alternative, supporting open-ended semantic category perception. Current VLMs and VFMs are limited to specific 2D single-image tasks, such as captioning and segmentation.
These methods often struggle with multimodal interactions and multi-view consistency, potentially leading to mismatches and 3D semantic ambiguities. Moreover, they lack a comprehensive understanding of the entire 3D space. To overcome these limitations, we propose a guidance framework centered around semantic attention maps and resolve ambiguities through scene reconstruction, thereby preserving 3D semantic and geometric coherence.

\vspace{-4mm}
\paragraph{Semantic Attention Map.} 
We employ semantic attention maps to integrate and guide the acquisition of desired prior knowledge from vision-language models at the semantic level. Given a multi-view image sequence, we prompt the VLM~\cite{chen2024internvl} to consistently generate all possible object categories within each image. Specifically, we use the attention map generation method~\cite{abnar2020quantifying, lin2024training} to compute and aggregate the attentions from transformer decoder, with $N$ output tokens $ S = s_1, \cdots, s_N $ and the attention tensor $ A \in \mathbb{R}^{H \times L \times N \times N} $, with $H$ attention heads, $L$ layers:
\begin{equation}
\begin{aligned}
    Attn(s_n^{l}) = \sum_{l=0}^{L} (\frac{1}{|H^{\prime}|} \sum_{h \in H^{\prime}}A_{h,l,k,j}) ,
\end{aligned}
\end{equation}
where $s_n^l \in S$ is the output of $n$-th semantic from the transformer layer $l \in L$, $A_{h,l,k,j}$ is the attention tensor between query $j$ and key $k$ in the head $h$ among subset of heads $H^{\prime}$.
We then rasterize the attention maps corresponding to these semantic categories into 2D feature maps, with each category represented by an aggregated attention map $M$. Notably, we establish a dynamically updated query list that incorporates the semantic information generated by VLMs.
We implement a semantic integration strategy that merges similar sub-vocabularies with excessive gradients into unified semantic categories, thereby enhancing efficiency and mitigating visual ambiguity. For instance, we consolidate ``tree'' and ``shrub'' under the general term ``vegetation''.

\vspace{-4mm}
\paragraph{Attention-guided Visual Prior.} 
Semantic attention maps unveil category-related visual cues, which we subsequently leverage to guide the generation of semantic-aligned masks and depth information. Concretely, we input semantic attention maps as prompt cues into the off-the-shelf segmentation models~\cite{zhao2023fast, zhang2023faster}, which then generate multiple masks within the region of interest. These masks are merged into instance-level candidate masks to fully delineate the targeted semantic regions. The mask with the highest similarity score to the embeddings of the semantic attention query is then selected.

In parallel, we employ semantic attention maps to guide depth estimation~\cite{piccinelli2024unidepth, yang2024depth} at the semantic level, decoupling background and foreground objects while excluding sky regions to avoid interference from infinite distances. We then aggregate depth information from multi-view images using semantic attention cues, where pixels within each area of interest yield a set of pseudo 3D point clouds that represent an individual instance.

\subsection{VL-GS}
Although vision-language guidance provides valuable priors, 3D occupancy annotation still encounters three major challenges: 1) Semantic conflicts across multiple views make naïve 2D-to-3D projection prone to misalignment and ambiguity; 2) Errors in geometry and depth estimation can lead to distortions and misplacements in 3D space; 3) Dynamic objects disrupt both spatial and temporal consistency in semantics and geometry.

To overcome these challenges, we propose Vision-Language Guided Gaussian Splatting (VL-GS), which efficiently reconstructs the entire scene while maintaining both semantic and geometric 3D consistency by combining attention-based priors and differentiable rendering. The core of VL-GS is the semantic-aware scalable GS, guided by semantic attention maps from vision-language models. During reconstruction, VL-GS smooths out 2D semantic ambiguities at the instance level and optimizes the geometric details of objects. We also introduce a self-estimated flow module to capture and reconstruct dynamic objects using temporally-aware dynamic Gaussians. 3D Semantic occupancy is then directly annotated through cumulative GS-Voxel splatting, which is both efficient and precise.

\begin{figure}[ht]
  \centering
  \includegraphics[width=1.0\linewidth]{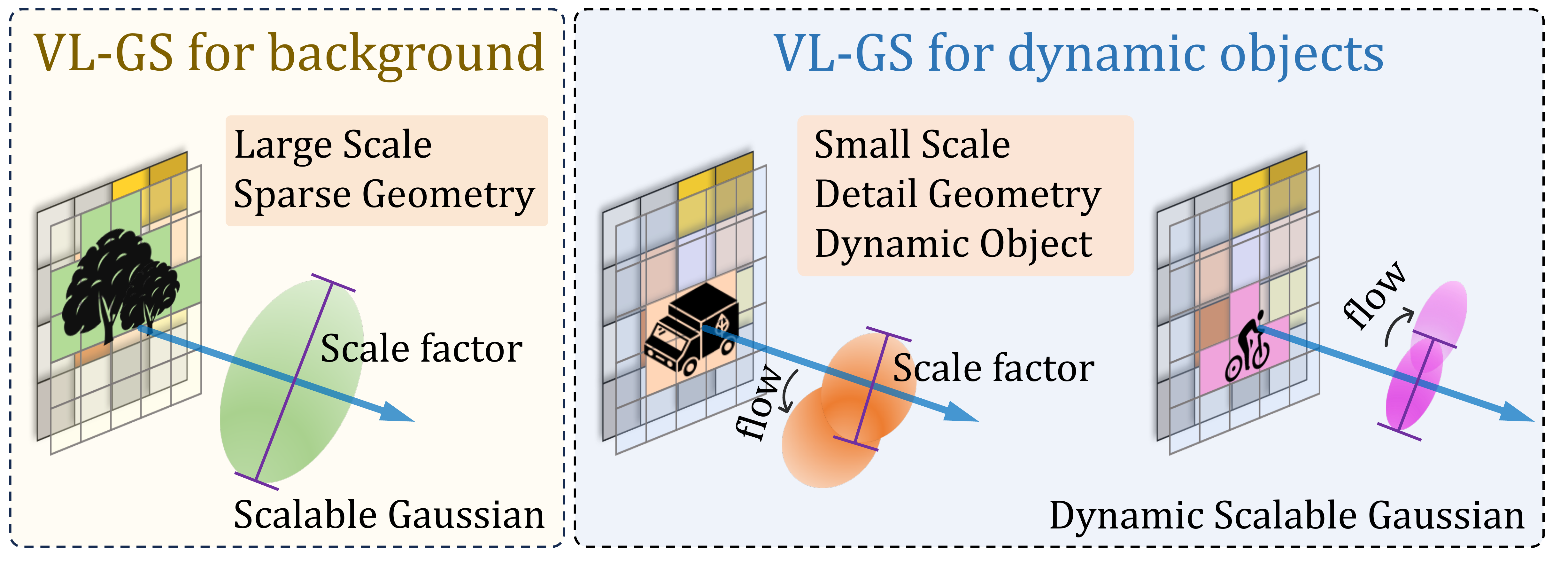}
  \vspace{-5mm}
  \caption{\textbf{Vision-Language Guided Gaussian Splatting (VL-GS)} efficiently reconstructs semantic instances using a scalable strategy guided by semantic attention maps from VLMs. Additionally, VL-GS models dynamic objects through dynamic Gaussians driven by self-estimated flow.}
  \label{fig:gaussians}
  \vspace{-4mm}
\end{figure}

\vspace{-4mm}
\paragraph{Semantic-aware Scalable Gaussian.} 
Obviously, different semantic objects occupy varying "weights" within a scene, which is intuitively reflected in their semantic occupancy across scales. Meanwhile, the ability to model at multiple granularities is expected to represent the diverse geometric complexities of instances. Based on this, we propose designing a semantic-aware scalable Gaussian that adaptively scales and reconstructs different semantic objects. Unlike dense voxels or point clouds, our method allows for representing regions of interest with sparse Gaussians, aided by scalability and semantic attention maps.

Given semantic attention cues from VLMs, we assign semantic attributes and corresponding scaling factors to each Gaussian. The blended semantic category of Gaussians can be obtained via $\alpha$-blending:
\begin{equation}
\begin{aligned}
    \Gamma_i = \sum_{i=1}^{N} \mathrm{softmax}(\gamma_i) \alpha_{i} \prod_{j=1}^{i-1}(1-\alpha_{j}) ,
\end{aligned}
\end{equation}
where $\Gamma_i$ is the rendered semantic for each pixel, weighted by the Gaussians' semantic attributes $\gamma$ and opacity $\alpha$. 
The scaling factor needs to be linearly related to the space occupied by each Gaussian, which cannot be simply calculated from the Gaussian centroid position $\{o_x,o_y,o_z\}$ due to the variations in anisotropic shape and spatial overlap. Thus, we estimate the occupied range of each Gaussian by considering the distance from the nearest tangent surface of the Gaussian ellipsoid to the voxel as:
\begin{equation}
\begin{aligned}
    d = o_z - \cfrac{\eta^{-1}\Sigma_{0,2}^{-1}(o_x - \kappa_x) + \eta^{-1}\Sigma_{1,2}^{-1}(o_y - \kappa_y)}{\Sigma_{2,2}^{-1}} ,
\end{aligned}
\end{equation}
where $d$ is the occupied depth from the voxel to the Gaussian ellipsoid, $\eta$ is the ray direction from the voxel center $k =(\kappa_x, \kappa_y, \kappa_z)$ to the Gaussian. $\Sigma$ is the covariance matrix, with $\Sigma_{i,j}$ denoting the corresponding matrix elements.
The Gaussian value $G(x)$ can be formulated as:
\begin{equation}
\begin{aligned}
    G(x) = e^{-\frac{1}{2}(\kappa-o)^{\top}\Sigma^{-1}(\kappa-o)} .
\end{aligned}
\end{equation}
The scaling factor is then adaptively adjusted based on the gradients of Gaussian values and the occupied range of the Gaussians. Notably, Gaussians of the same semantic category share similar scaling factor ranges, as objects with the same semantics exhibit comparable scales and geometries. As shown in Figure~\ref{fig:gaussians}, semantic-aware scalable gaussians enable the representation of large background areas (e.g., buildings) with sparse gaussians at a larger scale, while capturing finer geometries (e.g., cyclists) with denser gaussians at a smaller scale.

\vspace{-4mm}
\paragraph{Self-estimated Flow for Dynamic Objects.}
Dynamic objects could cause trailing effects due to temporal variations, thereby reducing the accuracy of occupancy annotation. Independently handling dynamic objects facilitates the enhancement of temporal and spatial consistency in semantics. Thus, we introduce a self-estimated 3D flow module, which is used to capture and aggregate dynamic objects. We also assign dynamic attributes to dynamic Gaussians to better model the motion of objects.

Specifically, we model the translation of each Gaussian kernel $p$ from time $t$ to time $t+ \Delta t$ as a flow vector $f$. Our goal is to minimize the point distances between the object's source points $U_1$ and target points $U_2$ to estimate the flow by applying Chamfer distance (CD)~\cite{fan2017point}. Since the same dynamic object is often represented by spatially adjacent Gaussians with the same semantics, we search for correspondences between paired points among the nearest Gaussian neighbors that share the same semantic:
\begin{equation}
\begin{aligned}
    CD(p,p^{\prime}) = \sum_{p \in U_1}\mathop{\min}_{p^{\prime} \in U_2}||p-p^{\prime} ||_2^2 + \sum_{p^{\prime} \in U_2}\mathop{\min}_{p \in U_1}||p^{\prime} -p||_2^2 ,
\end{aligned}
\end{equation}
where $p$ and $p^{\prime}$ are the positions of Gaussian kernels with the same semantic at time $t$ and $t+\Delta t$, respectively.
We define a dynamic indicator function between paired Gaussians to determine whether an object is in motion:
\begin{equation}
\begin{aligned}
    \mathbbm{1}(D) = \rho - \frac{1}{m} \sum_{i=1}^{m} \|p_{t+\Delta t}^{i} - p_t^{j}\|_{2} ,
\end{aligned}
\end{equation}
where $D$ is the average distance between paired Gaussians with the same semantic, $\rho$ is the dynamic threshold, and $m$ denotes the number of Gaussian ellipsoids. The centroid position at the $i$-th frame is denoted by $o_i$. Subsequently, we aggregate all temporally paired Gaussians based on the semantic attention map and motion cues.

\vspace{-4mm}
\paragraph{Geometric constraints from LiDAR.}
LiDAR points are widely used by existing occupancy annotation methods due to their precise geometric priors. Our pipeline also supports the use of LiDAR to obtain geometric constraints and continuously optimize the distribution of Gaussians.

Similar to~\cite{zhao2020fusion, zhao2023lif}, a point $p_{i,t}$ in the LiDAR sweep $L_t$ is projected onto the frame $I_t$, and its initial semantic label can be obtained by $K^{-1}[R^\top \phi_{x,y,t} + T]$, where $(K, R, T)$ are the corresponding camera parameters and homogenous transformation matrix, and $\phi$ is the pixel-level semantic label.
We aggregate the multi-frame of LiDAR points over time and compute the anchor centers $p_c = (x_c^i, y_c^i, z_c^i)$. We then implement a geometry-aware loss to enforce the alignment of Gaussian ellipsoid distributions with the geometric priors of their corresponding semantic regions:
\begin{equation}
\begin{aligned}
    L_{geo} = -\sum_{c=1}^{C} \sum_{i=1}^{M} \frac{1}{\| o_c(i) - p_c(i) \|_{2}^{2}} ,
\end{aligned}
\end{equation}
where $C$ denotes the number of semantic categories, $M$ is the number of Gaussian ellipsoid centers within the anchor range, and $o_i$ is the coordinate of the $i$-th Gaussian center.

\vspace{-4mm}
\paragraph{Cumulative GS-Voxel Splatting.}
Finally, we cumulatively splat VL-GS onto the voxel grid at an arbitrary voxel size, with each voxel's semantic label determined by weighting the occupied range and opacity from Gaussians:
\begin{equation}
\begin{aligned}
    \digamma(o) = \sum_{i=1}^{N} d_i G(x_i) \alpha_{i} \mathrm{softmax}(\gamma_i) ,
\end{aligned}
\end{equation}
where $d_i$ is the occupied depth of the Gaussian-to-3D voxel, treated as the splatting weight coefficient. $\alpha_{i}$ is the opacity, and $\mathrm{softmax}(\gamma_i)$ computes the semantic probability.

\begin{table*}[t]
\footnotesize
\setlength{\tabcolsep}{0.004\linewidth}
\centering
\caption{
\textbf{Semantic occupancy annotation on Occ3D-nuScenes~\cite{tian2024occ3d}.} C represents camera, and L denotes LiDAR. ``cons. veh.'' and ``drive. surf.'' stand for construction vehicles and driveable surfaces, respectively.
\ourmethod{}-V uses only images as input, while \ourmethod{}-M integrates both camera and LiDAR data. The intersection over union (IoU)
and mean IoU of semantic classes (mIoU) are calculated over all voxels. For fair comparisions, we replicate SurroundOcc*~\cite{wei2023surroundocc} and  OpenOcc*~\cite{Wang_2023_ICCV} by replacing the manually annotated results with the semantic point clouds projected from VLMs.
} 
\vspace{-3mm}
\begin{tabular}{l | c c c | c c c c c c c c c c c c c c c}

    \toprule
    Method 
    & \rotatebox{90}{Input}
    & \rotatebox{90}{IoU $\uparrow$}
    & \rotatebox{90}{mIoU $\uparrow$}
    & \rotatebox{90}{\textcolor{nbarrier}{$\blacksquare$} barrier} %
    & \rotatebox{90}{\textcolor{nbicycle}{$\blacksquare$} bicycle} %
    & \rotatebox{90}{\textcolor{nbus}{$\blacksquare$} bus} %
    & \rotatebox{90}{\textcolor{ncar}{$\blacksquare$} car} %
    & \rotatebox{90}{\textcolor{nconstruct}{$\blacksquare$} cons. veh.} %
    & \rotatebox{90}{\textcolor{nmotor}{$\blacksquare$} motorcycle} %
    & \rotatebox{90}{\textcolor{npedestrian}{$\blacksquare$} pedestrian} %
    & \rotatebox{90}{\textcolor{ntraffic}{$\blacksquare$} traffic cone} %
    & \rotatebox{90}{\textcolor{ntrailer}{$\blacksquare$} trailer} %
    & \rotatebox{90}{\textcolor{ntruck}{$\blacksquare$} truck} %
    & \rotatebox{90}{\textcolor{ndriveable}{$\blacksquare$} drive. surf.} %
    & \rotatebox{90}{\textcolor{nsidewalk}{$\blacksquare$} sidewalk} %
    & \rotatebox{90}{\textcolor{nterrain}{$\blacksquare$} terrain} %
    & \rotatebox{90}{\textcolor{nmanmade}{$\blacksquare$} manmade} %
    & \rotatebox{90}{\textcolor{nvegetation}{$\blacksquare$} vegetation} \\ %
    \midrule

\rowcolor{LGray} GaussianOcc~\cite{wan2024gaussianocc}  & C & 51.22 & 12.59 & 1.88 & 6.42 & 13.94 & 16.75 & 2.02 & 3.41 & 6.84 & 12.33 & 1.75 & 10.32 & 41.28 & 19.32 & 18.26 & 12.41 & 21.88 \\
LangOcc~\cite{boeder2024langocc}  & C & 46.55 & 12.04 & 2.73 & 7.21 & 5.78 & 13.92 & 0.51 & 10.80 & 6.42 & 8.67 & 3.24 & 11.02 & 42.10 & 12.44 & 27.17 & 14.13 & 14.55 \\
\rowcolor{LGray} VEON~\cite{zheng2025veon}  & C & 57.92 & 14.51 & 5.03 & 4.65 & 13.88 & 11.04 & 9.63 & 10.25 & 4.51 & 10.99 & 4.32 & 12.63 & 47.50 & 11.43 & 20.52 & 25.43 & 25.76 \\
SurroundOcc*~\cite{wei2023surroundocc}  & L & 68.87 & 18.59 & 18.68 & 17.23 & 18.19 & 18.31 & 10.27 & 18.29 & 17.34 & 14.95 & 21.19 & 19.88 & 21.33 & 20.74 & 18.11 & 23.26 & 21.02 \\
\rowcolor{LGray} OpenOcc*~\cite{Wang_2023_ICCV}  & C\&L & 70.59 & 17.76 & 23.73 & 8.06 & 26.10 & 22.95 & 11.72 & 11.59 & 10.36 & 9.72 & 5.60 & 19.13 & 39.51 & 22.15 & 20.87 & 13.19 & 21.81 \\
VLM-LiDAR  & C\&L & 73.28 & 16.32 & 13.34 & 10.37 & 17.04 & 20.65 & 7.26 & 15.20 & 14.61 & 5.88 & 19.40 & 21.47 & 15.13 & 13.32 & 15.74 & 28.17 & 27.24 \\
\rowcolor{LGray} OVIR-3D~\cite{lu2023ovir}  & C\&L & 54.30 & 18.47 & 18.54 & 10.69 & 15.30 & 23.82 & 9.42 & 13.13 & 11.57 & 8.32 & 10.19 & 20.49 & 36.85 & 24.22 & 21.84 & 16.30 & 36.33 \\
\rowcolor{violet!10} \textbf{\ourmethod{}-V}  & C & \underline{83.01} & \underline{20.92} & 12.70 & 10.45 & 7.81 & 20.42 & 5.79 & 17.58 & 18.50 & 24.25 & 4.23 & 12.88 & 55.54 & 24.23 & 27.14 & 35.62 & 36.61 \\
\rowcolor{violet!10} \textbf{\ourmethod{}-M}  & C\&L & \underline{88.62} & \underline{25.84} & 21.19 & 16.08 & 18.42 & 25.90 & 4.32 & 14.58 & 25.62 & 27.18 & 3.51 & 20.93 & 58.38 & 32.03 & 29.80 & 46.15 & 43.59 \\
    
\bottomrule
\end{tabular}
\vspace{-2mm}
\label{table:occ3d-nusc}
\end{table*}

\definecolor{nbarrier}{RGB}{255, 120, 50}
\definecolor{nbicycle}{RGB}{255, 192, 203}
\definecolor{nbus}{RGB}{255, 255, 0}
\definecolor{ncar}{RGB}{0, 150, 245}
\definecolor{nconstruct}{RGB}{0, 255, 255}
\definecolor{nmotor}{RGB}{200, 180, 0}
\definecolor{npedestrian}{RGB}{255, 0, 0}
\definecolor{ntraffic}{RGB}{255, 240, 150}
\definecolor{ntrailer}{RGB}{135, 60, 0}
\definecolor{ntruck}{RGB}{160, 32, 240}
\definecolor{ndriveable}{RGB}{255, 0, 255}
\definecolor{nother}{RGB}{139, 137, 137}
\definecolor{nsidewalk}{RGB}{75, 0, 75}
\definecolor{nterrain}{RGB}{150, 240, 80}
\definecolor{nmanmade}{RGB}{213, 213, 213}
\definecolor{nvegetation}{RGB}{0, 175, 0}
\definecolor{nfence}{RGB}{35, 135, 230}
\definecolor{ntrunk}{RGB}{195,85,85}
\definecolor{npole}{RGB}{213, 0, 139}

\section{Experiments}
\label{sec:expe}

\begin{table*}[t]
\footnotesize
\setlength{\tabcolsep}{0.002\linewidth}
\centering
\caption{
\textbf{Zero-shot cross-dataset performance on SemanticKITTI~\cite{behley2019semantickitti}.}
Novel class refers to entirely new, unseen semantics in nuScenes, while base class includes those seen during training. All compared methods are trained on Occ3D-nuScenes and evaluated on SemanticKITTI. Metric mIoU-base denotes the mIoU computed solely on base classes from Occ3D-nuScenes.
} 
\vspace{-1mm}
\begin{tabular}{l | c c c | c c c c c c c | c c c c c c c c c c c | c}

    \toprule
    \multicolumn{4}{c}{\textbf{(a) Val: SemanticKITTI}} & \multicolumn{7}{c}{\textbf{(b) Novel Class}} & \multicolumn{12}{c}{\textbf{(c) Base Class}}  \\
    \hline
    Method 
    & \rotatebox{90}{Input}
    & \rotatebox{90}{IoU $\uparrow$}
    & \rotatebox{90}{mIoU $\uparrow$}
    & \rotatebox{90}{\textcolor{ntrailer}{$\blacksquare$} bicyclist} %
    & \rotatebox{90}{\textcolor{nbus}{$\blacksquare$} moto-cyc.} %
    & \rotatebox{90}{\textcolor{nbarrier}{$\blacksquare$} parking} %
    & \rotatebox{90}{\textcolor{nfence}{$\blacksquare$} fence} %
    & \rotatebox{90}{\textcolor{ntrunk}{$\blacksquare$} trunk} %
    & \rotatebox{90}{\textcolor{npole}{$\blacksquare$} pole} %
    & \rotatebox{90}{\textcolor{ntraffic}{$\blacksquare$} traffic-sign} %
    & \rotatebox{90}{\textcolor{nbicycle}{$\blacksquare$} bicycle} %
    & \rotatebox{90}{\textcolor{ncar}{$\blacksquare$} car} %
    & \rotatebox{90}{\textcolor{nconstruct}{$\blacksquare$} other-veh.} %
    & \rotatebox{90}{\textcolor{nmotor}{$\blacksquare$} motorcycle} %
    & \rotatebox{90}{\textcolor{npedestrian}{$\blacksquare$} pedestrian} %
    & \rotatebox{90}{\textcolor{ntruck}{$\blacksquare$} truck} %
    & \rotatebox{90}{\textcolor{ndriveable}{$\blacksquare$} road} %
    & \rotatebox{90}{\textcolor{nsidewalk}{$\blacksquare$} sidewalk} %
    & \rotatebox{90}{\textcolor{nterrain}{$\blacksquare$} terrain} %
    & \rotatebox{90}{\textcolor{nmanmade}{$\blacksquare$} building} %
    & \rotatebox{90}{\textcolor{nvegetation}{$\blacksquare$} vegetation}
    & \rotatebox{90}{mIoU-base $\uparrow$}

    \\ %
    \midrule
    
\rowcolor{LGray} GaussianOcc~\cite{wan2024gaussianocc} & C & 22.42 & 4.18 & 0.0 & 0.0 & 0.0 & 0.0 & 0.0 & 0.0 & 0.0 & 1.33 & 7.10 & 2.81 & 3.06 & 2.91 & 3.42 & 15.80 & 10.43 & 3.78 & 2.55 & 22.11 & 6.84 \\
%
OVO~\cite{tan2023ovo} & C & 20.94 & 5.83 & 0.90 & 0.0 & 0.68 & 3.50 & 2.31 & 0.60 & 2.20 & 0.40 & 12.70 & 3.50 & 0.20 & 0.74 & 0.70 & 19.44 & 24.81 & 4.86 & 11.70 & 15.62 & 8.61 \\
\rowcolor{LGray} SurroundOcc~\cite{wei2023surroundocc} & L & 27.83 & 6.39 & 0.0 & 0.0 & 0.0 & 0.0 & 0.0 & 0.0 & 0.0 & 1.52 & 23.19 & 4.81 & 6.71 & 4.37 & 3.16 & 24.32 & 11.98 & 9.95 & 5.79 & 19.14 & 10.45 \\
VLM-LiDAR & C\&L & 28.12 & 5.32 & 0.0 & 0.0 & 0.0 & 0.0 & 0.0 & 0.0 & 0.0 & 2.04 & 19.17 & 3.31 & 2.13 & 2.64 & 5.89 & 19.02 & 16.58 & 6.31 & 3.59 & 14.98 & 8.69 \\
\rowcolor{violet!10} \textbf{\ourmethod{}-V}   & C & \underline{35.64} & \underline{9.36} & 1.38 & 3.60 & 0.59 & 4.34 & 5.36 & 14.32 & 6.62 & 4.71 & 22.29 & 3.89 & 10.35 & 7.54 & 8.78 & 26.14 & 15.66 & 9.84 & 4.14 & 18.87 & \underline{12.02} \\
\rowcolor{violet!10} \textbf{\ourmethod{}-M}   & C\&L & \underline{41.23} & \underline{12.76} & 1.27 & 5.23 & 0.33 & 5.71 & 5.97 & 15.17 & 8.72 & 7.83 & 24.60 & 4.92 & 9.30 & 11.18 & 8.39 & 44.74 & 24.43 & 5.85 & 17.01 & 29.12 & \underline{17.03} \\

\bottomrule
\end{tabular}
\label{table:semantic-kitti}
\end{table*}

 \begin{figure*}[ht]
  \centering
  \includegraphics[width=0.95\linewidth]{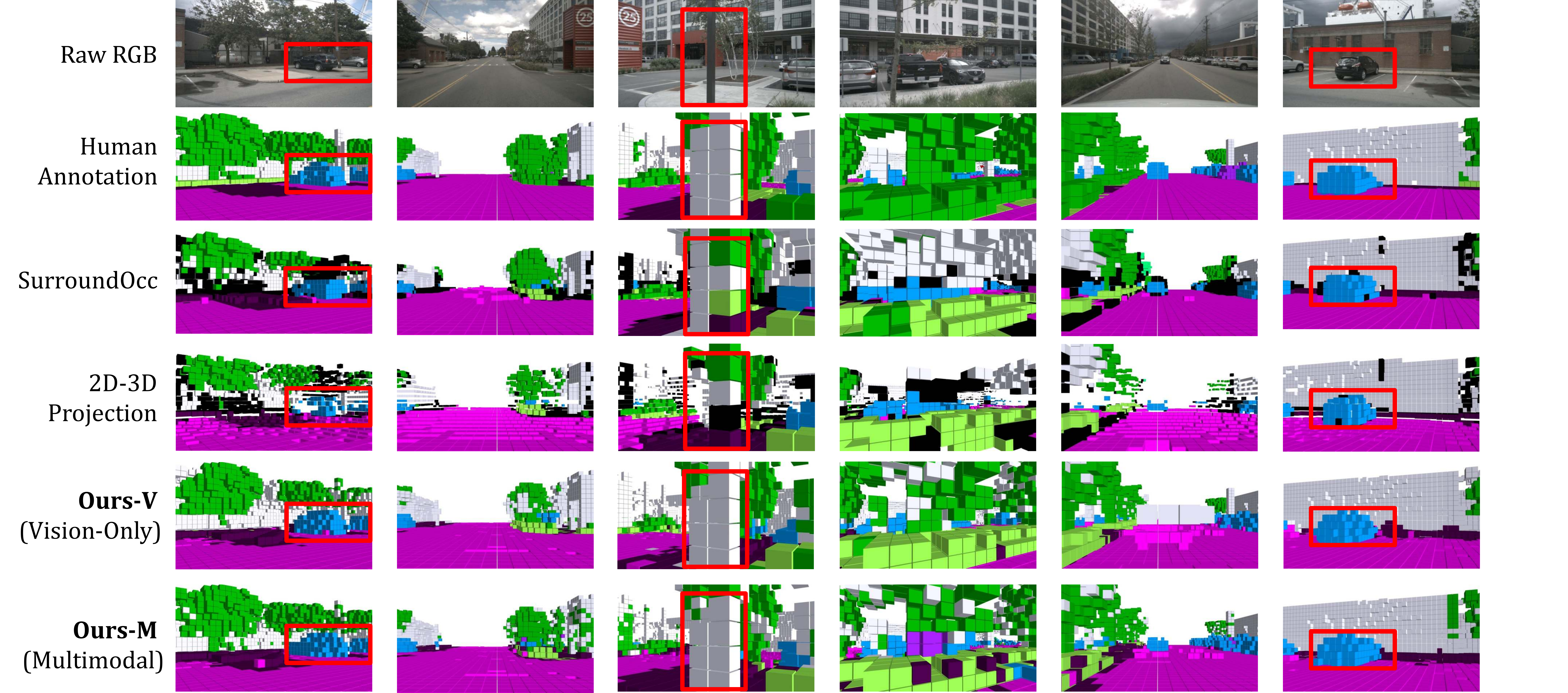}
  \vspace{-1mm}
  \caption{\textbf{Qualitative results of semantic occupancy annotation on Occ3D-nuScenes~\cite{tian2024occ3d}.} Our method enables high-quality annotation of semantic 3D occupancy, capturing fine-grained geometry, structurally challenging regions, and dynamic objects across complex scenes.}
  \label{figure:compare}
  \afterfig
 \end{figure*}

\subsection{Implementation Details} 
We use two benchmarks for evaluation: Occ3D-nuScenes, which is used to compare the performance of our method with other occupancy annotation methods for specific categories, while SemanticKITTI is used to assess the zero-shot capability across datasets and unseen categories.
We set the resolutions of images as 900 × 1600 for Occ3D-nuScenes and 370 × 1226 for SemanticKITTI. During optimization, we scale the image size to 225 × 400 and double it every 300 steps until reaching the original resolution.
We use the AdamW optimizer for optimization with an initial learning rate of 0.005. The learning rate for the position parameters decays every 250 steps with a decay rate of 0.98. Similar to~\cite{zhang2023occnerf, zheng2025veon}, we evaluate without the ``other'' and ``other flat'' classes. All experiments are carried out on 8 A100 GPUs.

\subsection{Performance Evaluation and Analysis}
We evaluate our method against the state-of-the-art (SOTA) methods for automatic semantic occupancy annotation, including offline methods~\cite{wei2023surroundocc, lu2023ovir, Wang_2023_ICCV} and self-supervised online methods~\cite{wan2024gaussianocc, boeder2024langocc, zheng2025veon}.

\vspace{-2mm}
\paragraph{Compared with point-based voxelization pipelines.}
Point-based voxelization annotation pipelines directly use LiDAR with 3D annotations (semantic points and 3D bounding boxes) as input. SurroundOcc~\cite{wei2023surroundocc} performs mesh reconstruction and a nearest neighbors algorithm to densify semantic points. OpenOcc~\cite{Wang_2023_ICCV} proposes the AAP pipeline to densify the voxel, followed by human post-processing to purify artifacts. For fair comparisons, we replicate these methods by replacing the manually annotated results with the semantic point clouds projected from VLMs. As shown in Table~\ref{table:occ3d-nusc}, our vision-centric method outperforms these pipelines that utilize LiDAR point clouds.

\paragraph{Compared with 2D-to-3D projection methods.}
Projecting annotated or generated 2D labels back onto a 3D representation is a natural idea, which is further refined by several methods~\cite{xu2023sampro3d, lu2023ovir}. However, these methods rely on pre-built 3D representations (e.g., point clouds or meshes) and employ multi-stage post-processing, including voting, filtering, and merging, to eliminate overlapping information. Undoubtedly, this strategy leads to the loss of crucial details and misalignment between semantics and representations. \ourmethod{} performs well against SAMPro3D~\cite{xu2023sampro3d} and OVIR-3D~\cite{lu2023ovir}, both of which project the outputs of SAM~\cite{ren2024grounded} onto 3D point clouds. 
We also design a baseline that directly projects the results of VLM and SAM onto LiDAR point clouds (VLM-LiDAR) and voxelizes them into semantic occupancy.
Table~\ref{table:occ3d-nusc} shows that \ourmethod{} still demonstrates better performance, based on the deep integration of VLM guidance and differentiable reconstruction.

\vspace{-3mm}
\paragraph{Compared with self-supervised methods.}
Self-supervised methods enable occupancy estimation from image features without relying on manual annotations. For a fair comparison, we extend existing self-supervised approaches by incorporating image sequences as historical frames and performing multi-frame feature aggregation. We further perform temporal fusion of the above outputs in the global coordinate system. As shown in Table~\ref{table:occ3d-nusc}, using pure visual input, our method outperforms GaussianOcc~\cite{wan2024gaussianocc}, which utilizes vanilla GS as an intermediate representation.
\ourmethod{} also performs well against LangOcc~\cite{boeder2024langocc} and VEON~\cite{zheng2025veon}, which are specifically designed for open-vocabulary occupancy estimation in surrounding-view scenes. 
While the aforementioned approaches do not require additional supervision, they struggle with efficiently modeling semantic geometry and neglect dynamic objects, leading to performance degradation.

\begin{figure}[ht]
  \centering
  \includegraphics[width=\linewidth]{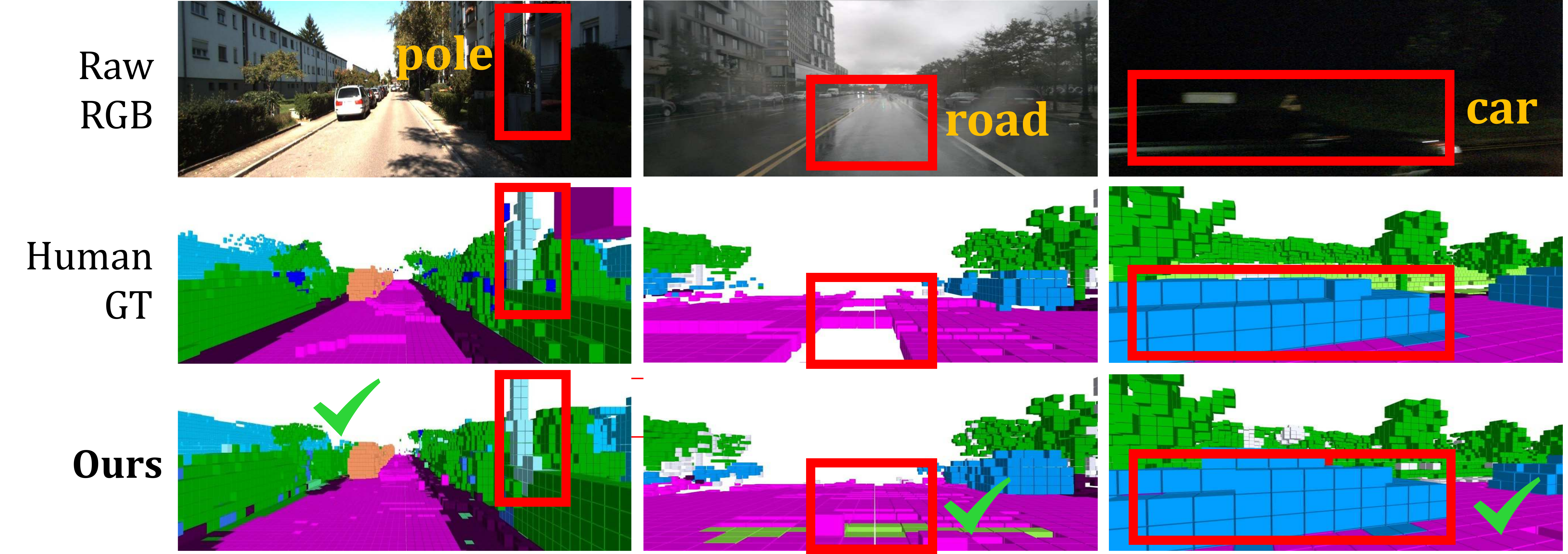}
  \caption{Qualitative comparison of our method with human annotations under complex lighting and extreme weather conditions.}
  \label{fig:day}
  \vspace{-4mm}
\end{figure}

\vspace{-3mm}
\paragraph{Qualitative results.} 
Figure~\ref{figure:compare} and~\ref{fig:day} show that our method excels in semantic occupancy annotation, showcasing superior scene completeness, consistency, and dynamic object handling, even without the use of LiDAR. In extreme weather conditions (e.g., rain and nighttime), our method maintains robust performance, achieving annotation results comparable to or even surpassing manually labeled ground truth. For instance, in areas where ground truth is missing due to rain, \ourmethod{} successfully reconstructs both the geometry and semantics of the road surface.

\subsection{Zero-shot and Generalization Ability}

SemanticKITTI differs from Occ3D-nuScenes in terms of semantic categories, sensor parameters, camera distribution, and voxel size. We evaluate on SemanticKITTI to verify the zero-shot and cross-dataset generalization capability.

\begin{table*}[ht]
\footnotesize
    \renewcommand\arraystretch{1.2}
    \setlength{\tabcolsep}{0.015\linewidth}
    \centering
    \caption{\textbf{Comparisons of annotation efficiency.} Open-ended stands for the annotation capability for undefined classes. Label-free means training without any human-labeled annotations. $\dagger$ indicates the use of VLMs to obtain 2D semantics instead of human labeling.
    }
    \label{tab:methods_comparison}
    \vspace{-2mm}
    \begin{tabular}{lcccccccccc}
    \Xhline{0.75pt}
    Method & Anno. Time & Input Modality & Representation & Memory &  Number & Open-Ended & Label-Free\\
        \hline

\rowcolor{LGray}    Auto+Human~\cite{Wang_2023_ICCV} & 4000+ human hours & L & Point Clouds & - & 1.2 M & \textcolor{red}{\usym{2717}} & \textcolor{red}{\usym{2717}} \\
    GaussianOcc~\cite{wan2024gaussianocc} $\dagger$ & $\approx$60 GPU hours & C & Vanilla GS & 32 G & 0.8 M & \textcolor{red}{\usym{2717}} & \textcolor{ForestGreen}{\usym{2713}}\\
\rowcolor{LGray}    \rowcolor{LGray} SurroundOcc~\cite{wei2023surroundocc} $\dagger$ & 1000+ GPU hours & L & Mesh \& Voxel & 73 G & 3.0 M & \textcolor{red}{\usym{2717}} & \textcolor{red}{\usym{2717}} \\
    VLM-LiDAR $\dagger$ & $\approx$50 GPU hours & C\&L & Point Clouds & 34 G & 1.2 M & \textcolor{red}{\usym{2717}}  & \textcolor{red}{\usym{2717}} \\
    \rowcolor{violet!10} \textbf{\ourmethod{}} $\dagger$ & $\approx$30 GPU hours & C or C\&L & VL-GS & 5.0 G & 0.3 M & \textcolor{ForestGreen}{\usym{2713}} & \textcolor{ForestGreen}{\usym{2713}} \\
        
    \Xhline{0.75pt}
    \end{tabular}
    \label{tab:efficiency}
    \vspace{-3mm}
\end{table*}

To evaluate the zero-shot and open-ended semantic annotation ability, we select novel classes from SemanticKITTI as the test set, which are not visible during the annotation process.
Table~\ref{table:semantic-kitti} shows that all self-supervised methods~\cite{ wan2024gaussianocc, tan2023ovo} suffer significant performance degradation, as they are tailored to specific camera parameters and occupancy distributions.
For novel classes unseen during learning, these methods fail to label undefined semantic occupancy.
Compared to offline annotation pipelines, including point-based voxelization and semantic projection, our method shows better robustness and enhanced capability for open-ended semantic annotation.

\subsection{Annotation Efficiency}
Table~\ref{tab:efficiency} presents evaluations on representation characteristics and model efficiency.
Notably, \ourmethod{} demonstrates an advantage in computational cost, delivering better performance with reduced memory requirements. 
In contrast, scene representations based on dense voxels and Point Clouds incur redundant computational costs.
In addition, \ourmethod{} strikes a balance between efficiency and flexibility, enabling open-ended scene-aware occupancy reconstruction, supporting open-vocabulary semantic occupancy annotation, and requiring no human-labeled annotations.

\begin{figure}[ht]
  \centering
  \includegraphics[width=1.0\linewidth]{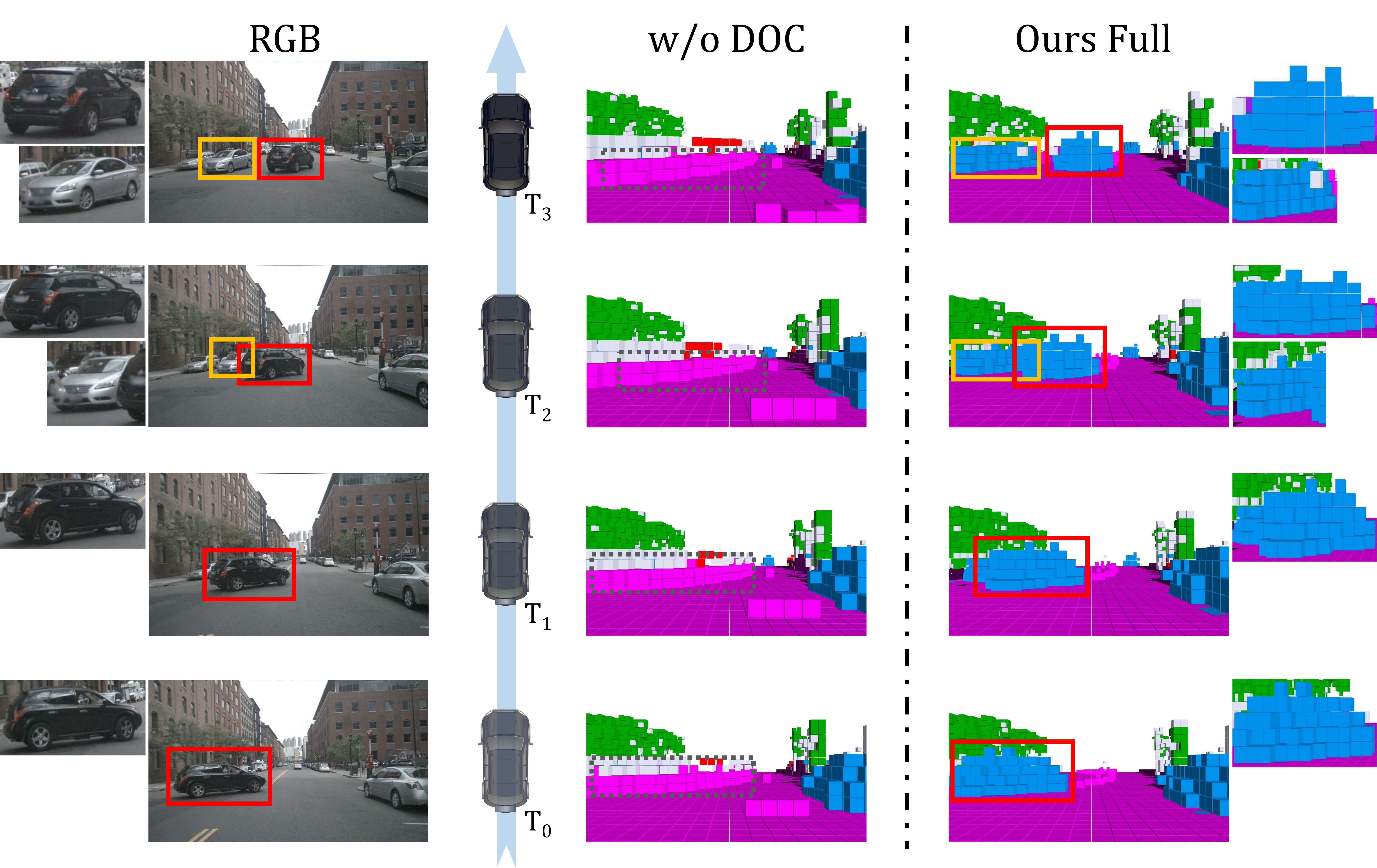}
  \caption{\textbf{Semantic occupancy of dynamics.} \ourmethod{} accurately annotates the semantic occupancy of dynamic objects, maintains spatiotemporal consistency, and infers occluded parts.}
  \label{fig:dynamic}
  \vspace{-2mm}
\end{figure}

\subsection{Ablation Studies}
We analyze the effect of the self-estimated flow module for dynamic objects by disabling the clustering of dynamic objects and optimizing them together with static foregrounds. Table~\ref{tab:Ablation} and Figure~\ref{fig:dynamic} show that the self-estimated flow module effectively mitigates the challenges of dynamic trailing and spatial occlusion in the annotation of occupancy.
We further ablate the effect of semantic-aware scalable Gaussians and LiDAR geometric priors by replacing the scalable Gaussians and removing the $L_{geo}$ loss from our framework. 
The degraded results highlight the importance of geometric priors in constraining the shape and distribution of Gaussians, leading to more accurate occupancy reconstruction.

\begin{table}[!t]
  \centering
  \caption{
  {\textbf{The effect of each module in our method.} SFM is short for the self-estimated flow module, and SSG denotes the employment of the semantic-aware scalable Gaussians.}}
    \footnotesize
  \centering
  \setlength{\tabcolsep}{8mm}{
  {
  \vspace{-2mm}
    \begin{tabular}{ccc}
 \toprule
    \textbf{Model} & \textbf{IoU $\uparrow$} & \textbf{mIoU $\uparrow$} \\
    \hline 
     w/o SFM             & 82.65     & 16.84  \\
     w/o SSG  &  80.27     &  17.67 \\
     w/o $L_{geo}$      &  81.49     &  20.36 \\
    \ourmethod{}-V               & 83.01 & 20.92  \\
     \ourmethod{}-M               &  88.62  & 25.84   \\
    \hline
    \end{tabular}%
    }
    }
  \aftertab
  \label{tab:Ablation}%
  \vspace{-3mm}
\end{table}%

\section{Conclusion}
In this paper, we propose \ourmethod{}, an vision-centric automated pipeline for open-ended semantic 3D occupancy annotation that integrates differentiable Gaussian splatting guided by vision-language models.
To facilitate scene understanding, we leverage VLMs and build an efficient and comprehensive scene representation for occupancy annotation. \ourmethod{} integrates vision-language attention with visual foundation models, effectively handles dynamic objects over time, and enhances both spatiotemporal consistency and geometric detail. 
Our framework achieves state-of-the-art performance on open-ended semantic occupancy annotation and performs favorably against other automated annotation pipeline, without using any human annotations.

\clearpage
\section*{Acknowledgment}
This work was supported by National Key R\&D Program of China (Grant No. 2022ZD0160305). This work was also a research achievement of Key Laboratory of Science, Technology, and Standard in Press Industry (Key Laboratory of Intelligent Press Media Technology). Ming-Hsuan Yang was supported in part by the Institute of Information \& Communications Technology Planning \& Evaluation (IITP) grant funded by the Korean Government (MSIT) (No. RS-2024-00457882, National AI Research Lab Project).

{
    \small
    \bibliographystyle{ieeenat_fullname}
    \bibliography{main}
}

\end{document}